\documentclass[]{article}
\usepackage[letterpaper]{geometry}
\usepackage{amta2022}
\usepackage{times}
\usepackage{url}
\usepackage{latexsym}
\usepackage{natbib}
\usepackage{layout}

\usepackage{booktabs}
\usepackage{multirow}
\usepackage{graphicx, subcaption}
\usepackage[flushleft]{threeparttable}
\usepackage{wrapfig}

\begin{document}

% \amtaHeader{x}{x}{xxx-xxx}{2015}{45-character paper description goes here}{Author(s) initials and last name go here}
\title{\bf AutoNMT: A Framework to Streamline the Research of Seq2Seq Models }

\author{\name{\bf Anonymous authors} \hfill  \addr{Paper under double-blind review} \\
}

\author{\name{\bf Salvador Carrión} \hfill  \addr{salcarpo@prhlt.upv.es}\\
        \name{\bf Francisco Casacuberta} \hfill \addr{fcn@prhlt.upv.es}\\
        \addr{PRHLT Research Center, Universitat Politècnica de València}
}

\maketitle
\pagestyle{empty}

\begin{abstract}

We present AutoNMT\footnote{\url{https://github.com/salvacarrion/autonmt}}, a framework to streamline the research of seq-to-seq models by automating the data pipeline (i.e., file management, data preprocessing, and exploratory analysis), automating experimentation in a toolkit-agnostic manner, which allows users to use either their own models or existing seq-to-seq toolkits such as Fairseq or OpenNMT, and finally, automating the report generation (plots and summaries). Furthermore, this library comes with its own seq-to-seq toolkit so that users can easily customize it for non-standard tasks.
\end{abstract}

\section{Introduction}

The performance of NMT models is often greatly affected by decisions such as the normalization used, the tokenization, the size of the vocabulary, the model, etc., decisions that must be remembered at both a training and inference time so that the model can perform as expected. In addition to this, it is also very common to train and evaluate these models on multiple datasets, which, added to the above, makes the research of seq-to-seq models a harder task than necessary.

Furthermore, in research, it is very common to prototype ideas rapidly and write throwaway code, which in complex pipelines can lead to small errors that can easily go unnoticed. In addition to this, researchers, more often than not, spend countless hours on time-consuming tasks that are not strictly related to their research, such as writing boilerplate code, debugging errors, creating charts, etc.

To address these challenges, we have built AutoNMT, a Python framework to make the research of seq-to-seq models an easier task and, therefore, allow researchers to spend more time on their ideas. This framework aims to automate as many tasks of the typical seq-to-seq pipeline as possible but without imposing further constraints on the researcher's workflow by:

\begin{itemize}
    \item Automating the data pipeline (i.e., file management, data preprocessing, and exploratory analysis).
    \item Automating the experimentation in a toolkit-agnostic environment so that users can use their own models, vocabularies, and toolkits (e.g., Fairseq, OpenNMT, HuggingFace, etc.).
    \item Managing reporting, logging, and versioning.
\end{itemize}

\section{Related Work}

In recent years, machine learning has enjoyed great popularity thanks to scientific advances in the field, which in many cases has ended up materialized into products such as Keras~\citep{keras}, Tensorflow~\citep{tensorflow}, PyTorch~\citep{pytorch}, HuggingFace~\citep{huggingface}, Scikit-learn~\citep{scikit-learn},... that make the lives of many scientists easier by allowing them to research more efficiently.

Under this premise, many products or libraries have appeared to streamline the workflow of engineers, researchers, and developers. For example, Fairseq~\citep{fairseq} and OpenNMT~\citep{opennmt} were solutions to deal with the complex training pipelines in Neural Machine Translation; HuggingFace~\citep{huggingface} on democratizing NLP; AutoML and Auto-Sklearn~\citep{automl} put their focus on automating the training and evaluation of specific machine learning models; Ray Tune~\citep{raytune} targeted the experiment execution and hyperparameter tuning at any scale; ONNX~\citep{onnx} was designed to solve the model interoperability; SentencePiece~\citep{sentencepiece} delivered an efficient implementation on many text tokenizers; etc.

Inspired by these libraries, we decided to go one step ahead and build a new tool to streamline the research of seq-to-seq models by building a library on top of these well-tested libraries: PyTorch~\citep{pytorch}, SentencePiece~\citep{sentencepiece}, Sacremoses~\citep{moses}, SacreBleu~\citep{sacrebleu}, Fairseq \citep{fairseq}, OpenNMT~\citep{opennmt}, among many others.

\section{AutoNMT Framework}

The core of this library is composed of three components: the \textit{Dataset Builder}, the \textit{Meta-Trainer}, and the utility to generate automatic reports.

\subsection{Dataset Builder}

\begin{figure}[ht]
\centering
\includegraphics[width=0.50\columnwidth]{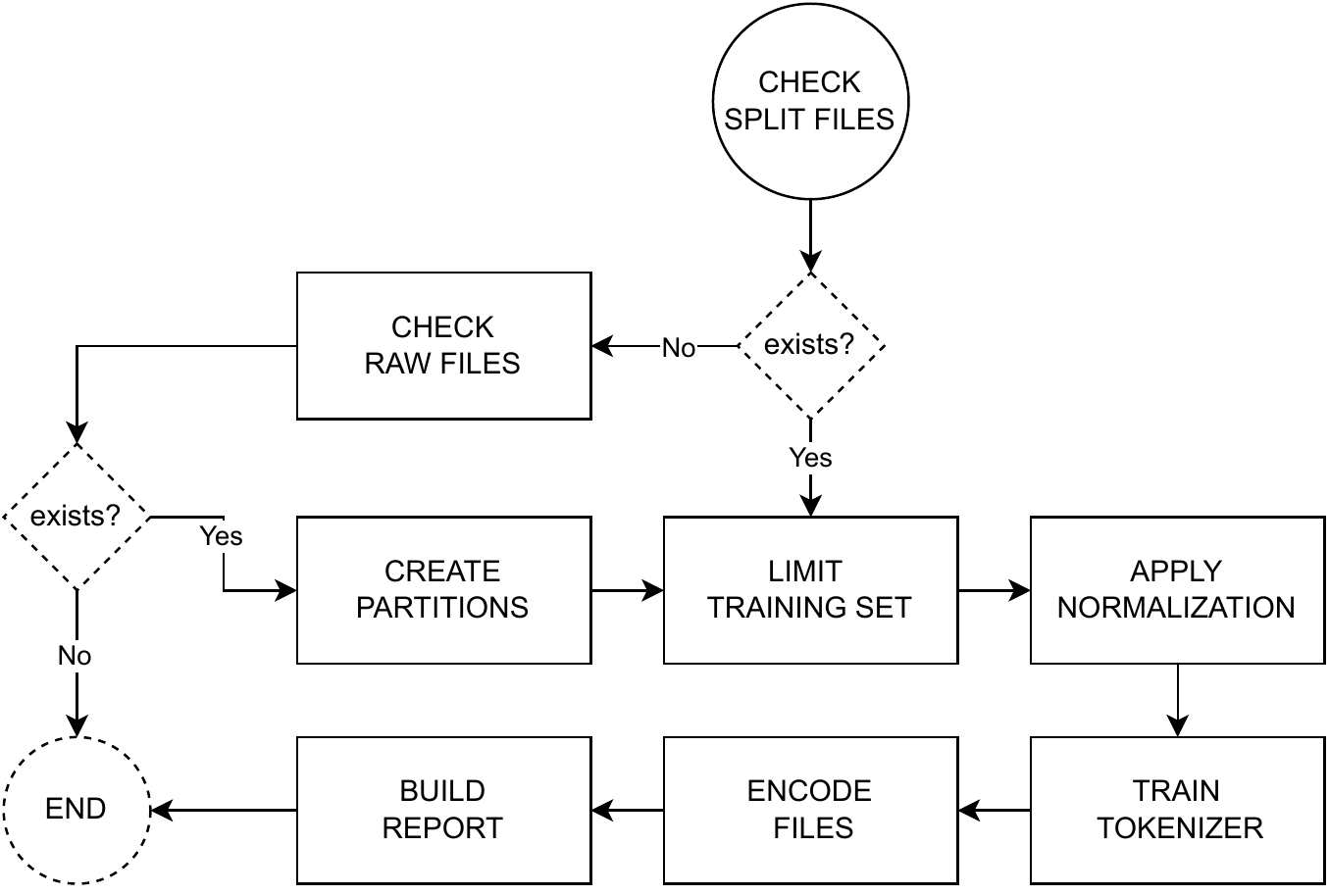}
    \caption{\textbf{Data pipeline}: Workflow of the \textit{DatasetBuilder} component}
    \label{fig:databuilder-diagram}
\end{figure}

The \textit{DatasetBuilder} is the class in charge of managing data and generating new datasets. First, it keeps organized the datasets, vocabularies, statistics, models, graphs, and reports. And second, it generates new datasets variants on-demand, using parameters such as:

\begin{itemize}
    \item \textbf{Normalizations}: NFD, NKFT, Strip, StripAccents, LowerCase, Replace,...
    \item \textbf{Tokenizations}: Bytes, Chars, Chars+Bytes, Unigram, Unigram+Bytes, BPE, BPE+Bytes, Words, Words+bytes and None\footnote{The \textit{None} option is convenient when the tokenization has to be done on the fly (e.g., data augmentation, subword regularization, etc.)}.
    \item \textbf{Vocabulary sizes}: List of vocabulary max. sizes (32K, 16K, 8K,...)
    \item \textbf{Training sizes}: Limits the number of sentences in the training set (10M, 1M, 100K,...)
\end{itemize}

The workflow of this component can be seen in Figure~\ref{fig:databuilder-diagram}. The data pipeline starts by checking if there are files to process (raw or splits); if no files are found, the component will ask the user\footnote{The interactive flag must be set to \textit{True} (default) } if it can create the directories where the user is expected to put the datasets so that later can be found.

\begin{figure}[ht]
\centering
\includegraphics[width=0.80\columnwidth]{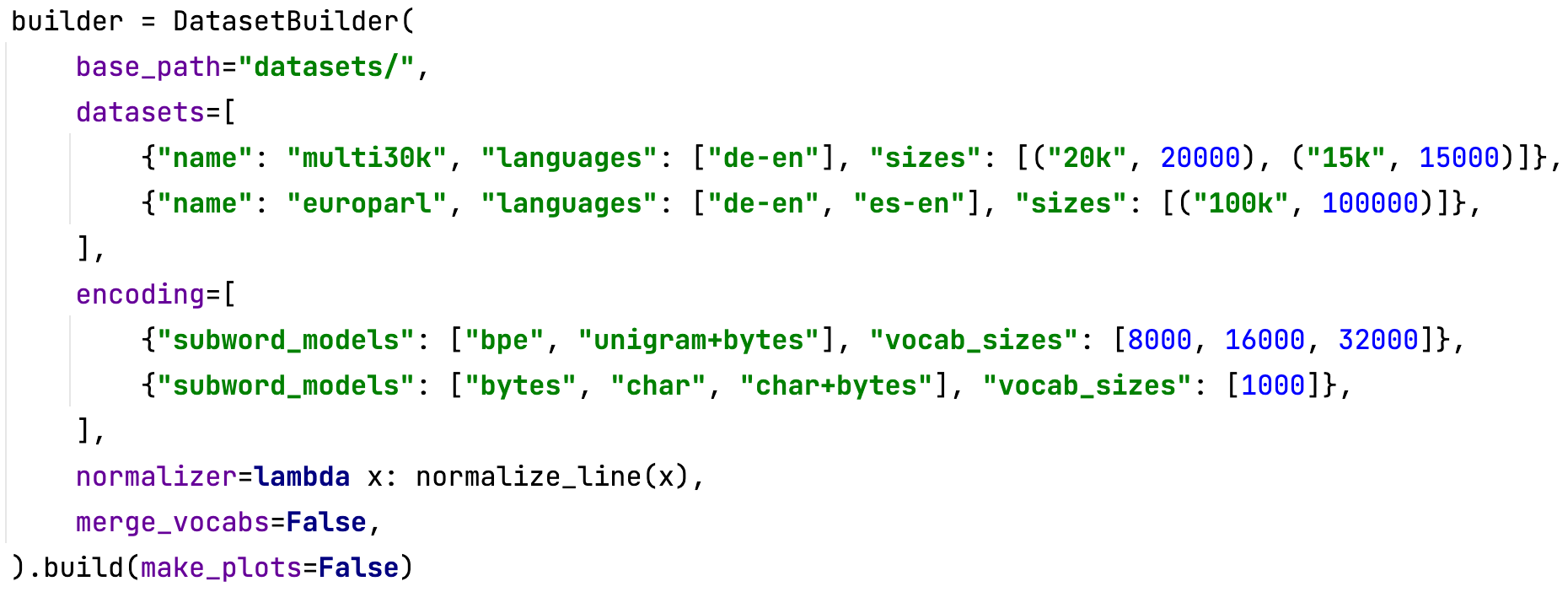}
    \caption{ \textbf{DatasetBuilder}. This code will create (1x2 + 2x1) datasets with (2x3 + 3x1) variations each to explore the effects of these settings in our models.}
    \label{fig:databuilder-code}
\end{figure}

To use this component, the user only needs to put the datasets (i.e., raw or split files) into their corresponding folders (these folders will be created interactively from the base path specified). After that, the \textit{DatasetBuilder} will be able to index and preprocess all datasets automatically. For example, in Figure~\ref{fig:databuilder-code} we can see the code for a DataBuilder instance that will generate a total of 36 datasets variants (18 for Multi30K\footnote{Multi30K: 1 language x 2 sizes (training set limit) x (2+3) subword models (with 3 and 1 vocab sizes)} and another 18 for Europarl\footnote{Europarl: 2 languages x 1 size (training set limit) x (2+3) subword models (with 3 and 1 vocab sizes)}, corresponding to different datasets, languages, sizes, tokenizations, and vocabularies.

\begin{figure}[ht]
\centering
\subfloat[Sentences per split \label{fig:dataset-stats-a}]
        {\includegraphics[width=0.25\columnwidth]{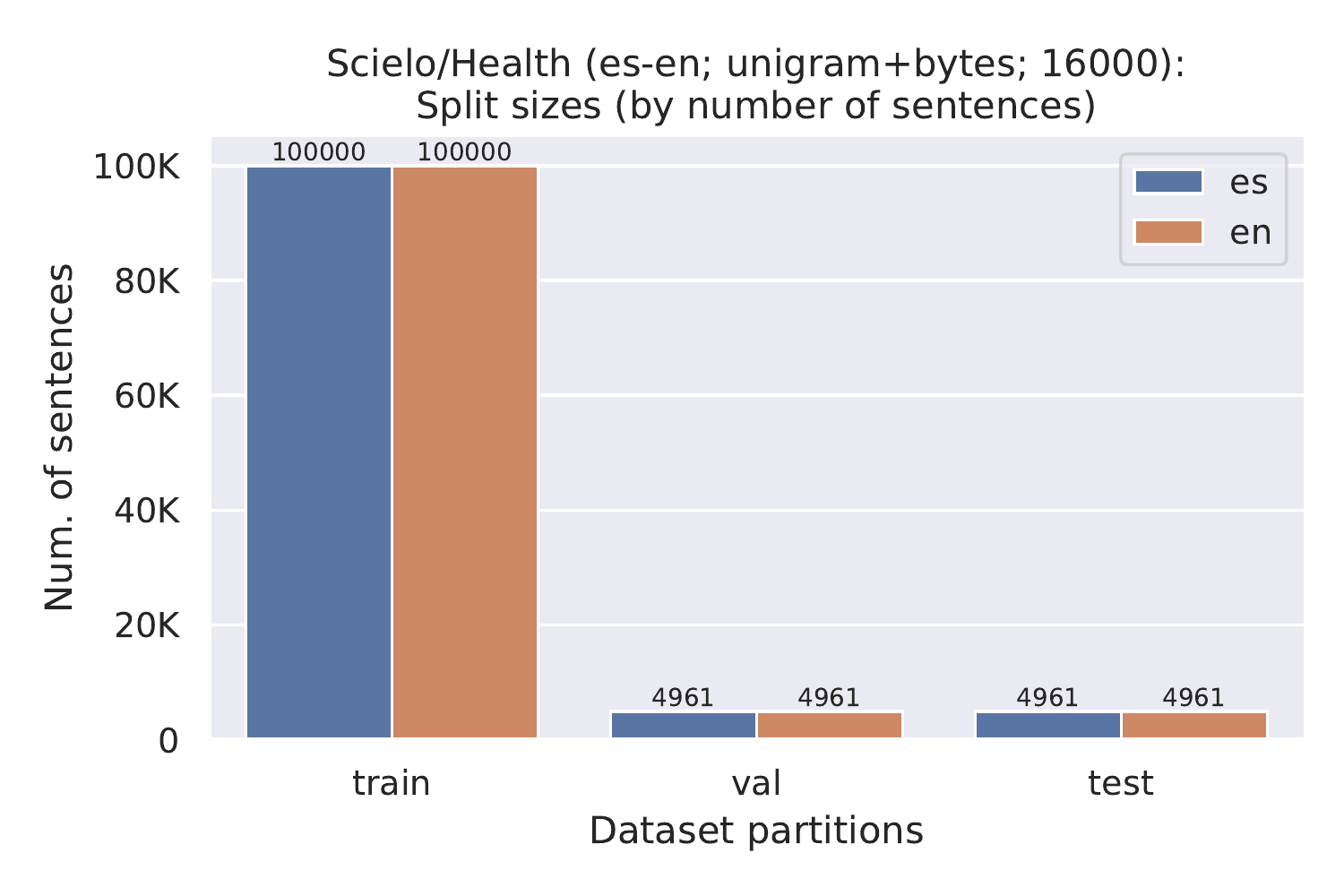}}
\subfloat[Tokens per split \label{fig:dataset-stats-a}]
    {\includegraphics[width=0.25\columnwidth]{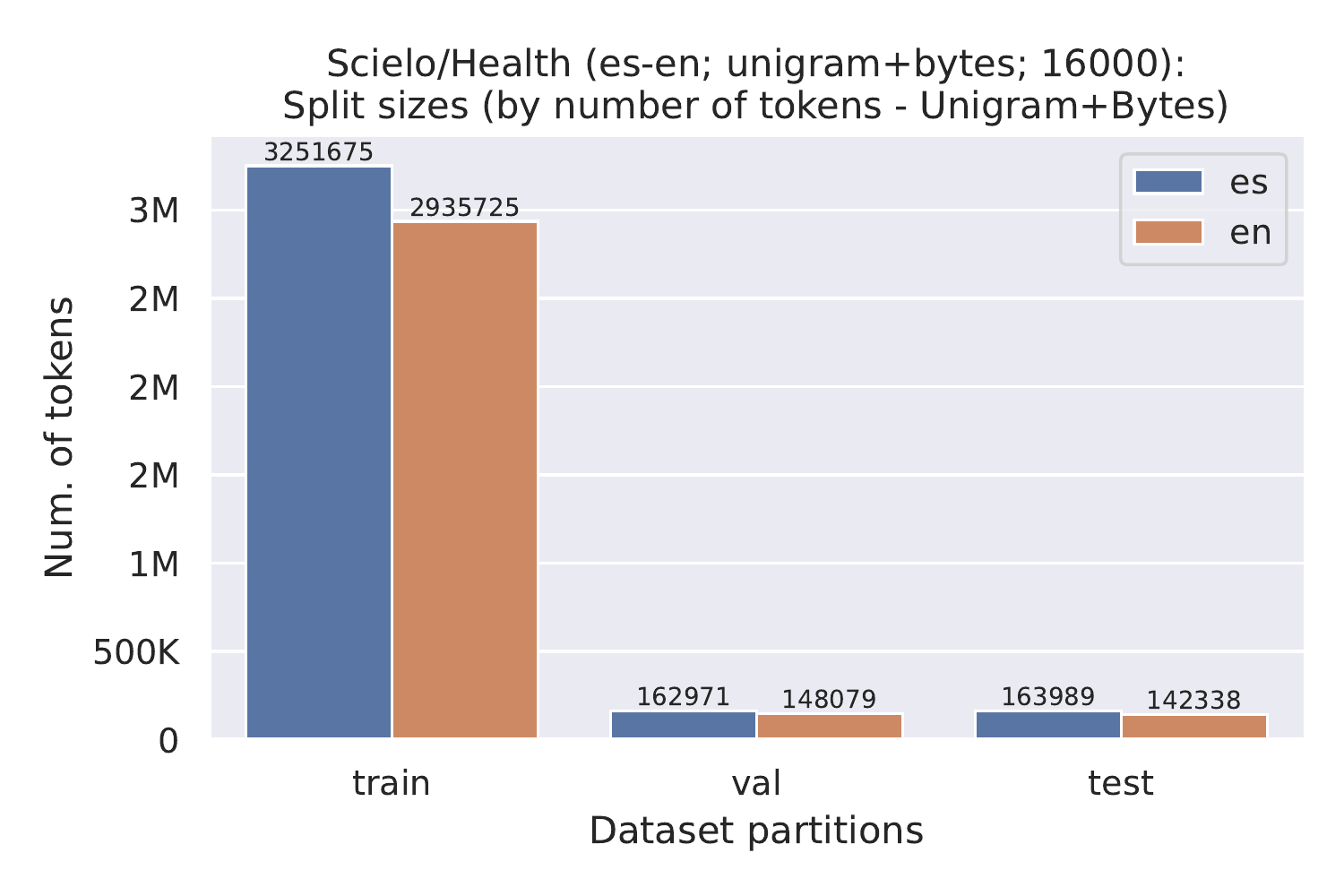}}
\subfloat[Length distribution \label{fig:dataset-stats-b}]
     {\includegraphics[width=0.25\columnwidth]{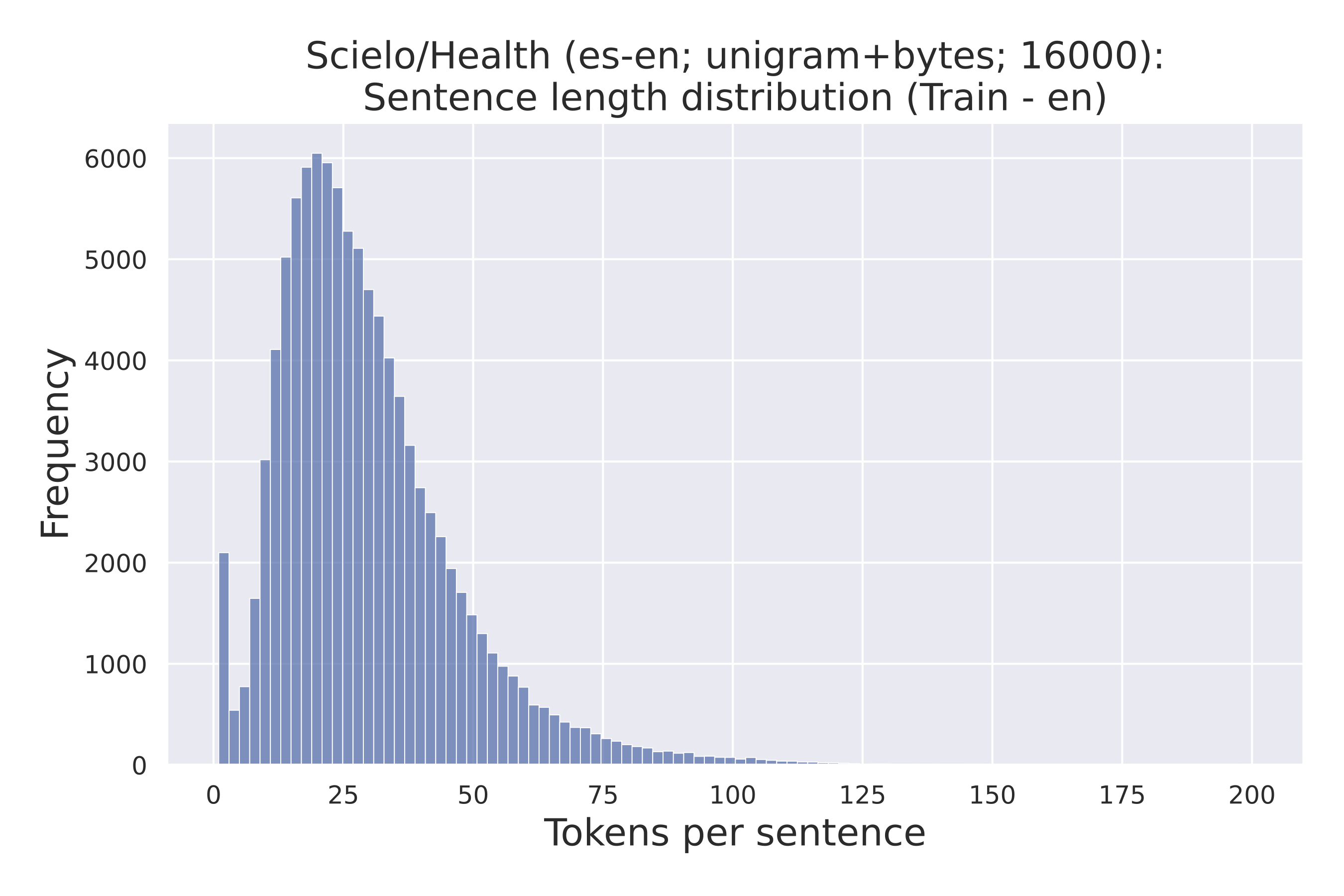}}
\subfloat[Token frequency \label{fig:dataset-stats-c}]
{\includegraphics[width=0.25\columnwidth]{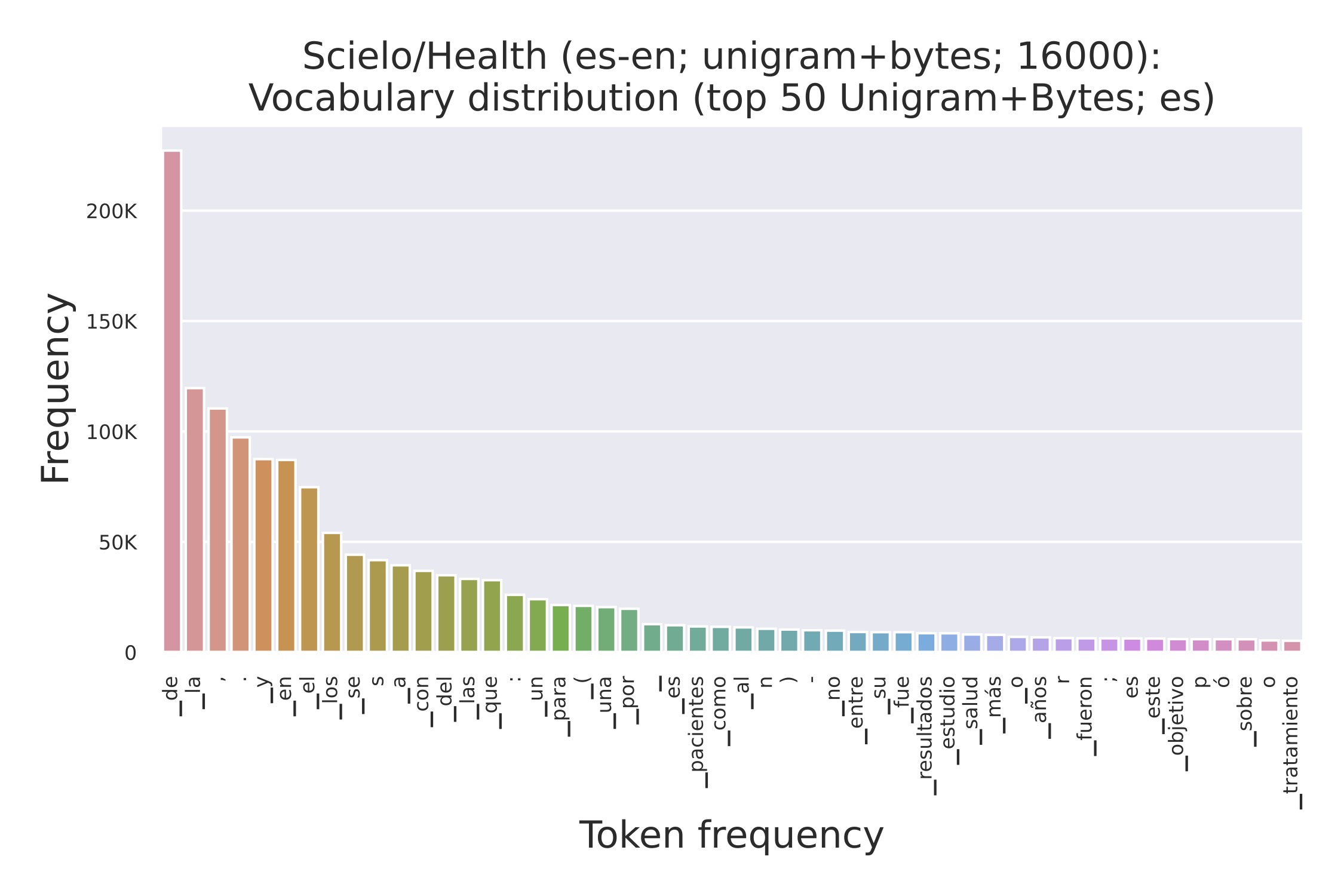}}
\caption{\textbf{Automatic Exploratory Analysis}: AutoNMT automatically generates basic reports, statistics and plots to describe the datasets and its vocabularies.}
\label{fig:dataset-stats}
\end{figure}

Furthermore, the \textit{DatasetBuilder} will handle part of the exploratory analysis by creating plots, stats, and reports that are typically used to describe the datasets, such as tokens per partition, sentence length distributions, token frequency, max/min/avg/ sentence length, unknowns per sentence, etc. (See Figure~\ref{fig:dataset-stats}).

\subsection{Meta-Trainer}

\begin{figure}[ht]
\centering
\includegraphics[width=0.65\columnwidth]{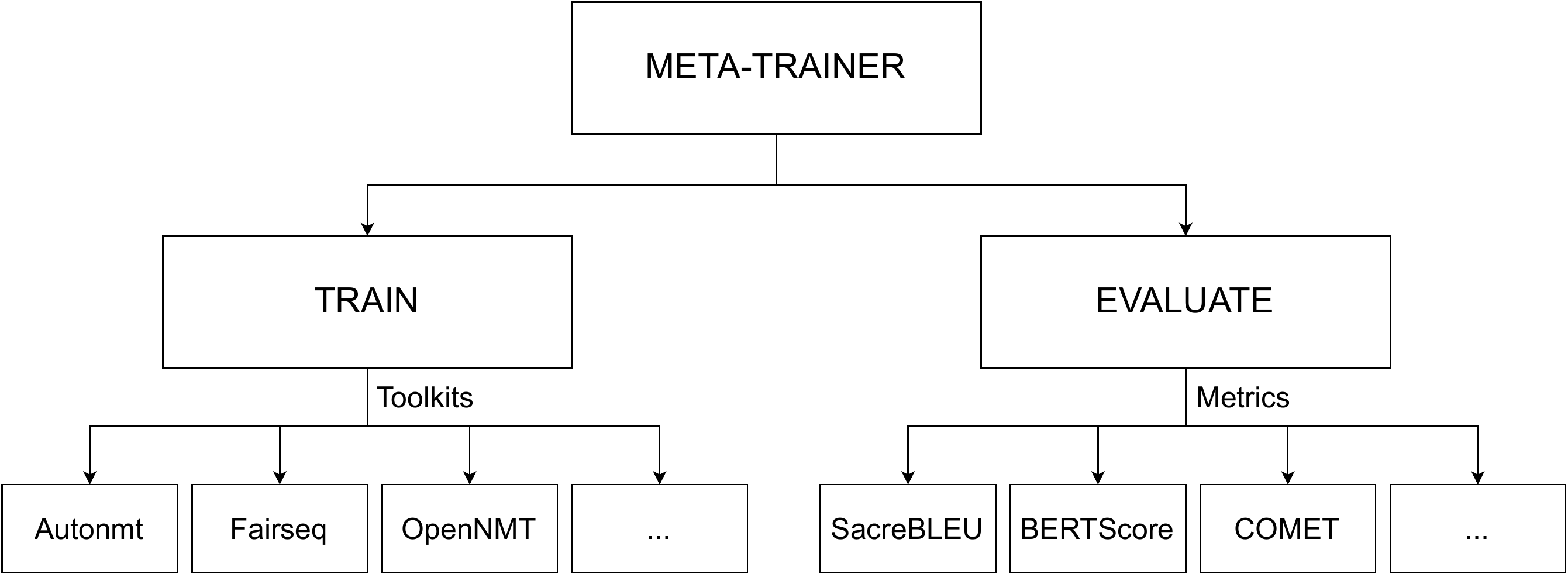}
    \caption{The \textbf{Meta-Trainer} has two functions: i) Training in a toolkit-agnostic manner; and ii) Evaluating the models against multiple metrics using a unified interface}
    \label{fig:meta-trainer-diagram}
\end{figure}

The Meta-Trainer has two functions: Training and Evaluation (See Figure~\ref{fig:meta-trainer-diagram}). First, it abstracts the seq-to-seq toolkit so that users can train their models using either their toolkit of preference (e.g., Fairseq, OpenNMT, AutoNMT) or a customized one that inherits from our toolkit. Second, it acts as a unified interface to evaluate each of the trained models against a large set of available metrics (BLEU~\citep{metric-bleu}, chrF~\citep{metric-chrf}, BERTScore~\citep{metric-bertscore}, COMET~\citep{metric-comet}, BEER~\citep{metric-beer}, etc.).

The motivation for building this meta-trainer was mainly due to three reasons:

\begin{itemize}
    \item First, to automate the training and evaluation of the models using the datasets generated by the DatasetBuilder without worrying about the inner workings of the data pipeline.
    \item Second, to automate the experimentation and, at the same time, allow users to use their preferred seq-to-seq toolkit through a minimal and unified interface.
    \item Third, to allow users to easily create customized models, trainings, and toolkits when the existing solutions could not meet their needs. Besides, they can compare their implementation against other toolkits in a controlled environment to ease its debugging.
\end{itemize}

\subsubsection{Training}

\begin{figure}[ht]
\centering
\subfloat[FairSeqTranslator \label{fig:toolkit-fairseq}]
    {\includegraphics[width=0.95\columnwidth]{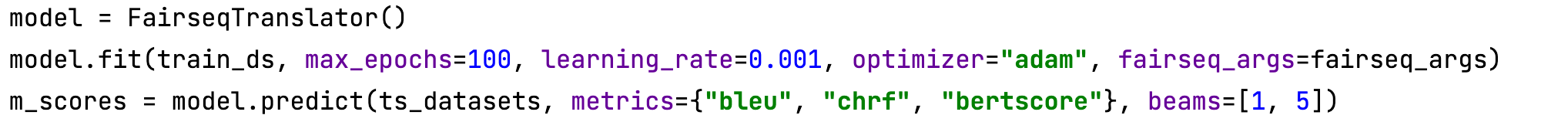}}
\hfill
\subfloat[AutonmtTranslator \label{fig:toolkit-autonmt}]
     {\includegraphics[width=0.95\columnwidth]{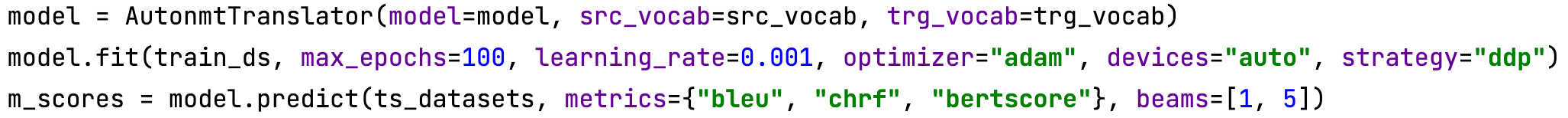}}
    \caption{ \textbf{Meta-Trainer}: This object abstracts the training component so that a user can train their models using our data pipeline with their preferred seq-to-seq toolkit. }
    \label{fig:toolkits-code}
\end{figure}

Since this object acts as a wrapper to automate the training in any of the supported toolkits, a user that is used to work with a toolkit such as FairSeq could simple instantiate the \textit{FairSeqTranslator} class and call its fit function to train a FairSeq model using the datasets generated by the \textit{DatasetBuilder} (See Figure~\ref{fig:toolkit-fairseq}). However, if a user wants to have more control over the training, the models, and its data, they could simply replace the FairSeqTranslator class with AutonmtTranslator class (See ~\ref{fig:toolkit-autonmt}). 

In the case that a user wanted to use a toolkit that is not supported, they would only have to create a new object that inherits from the \textit{BaseTranslator} class and override the \textit{preprocess}, \textit{train} and {translate} methods. On the other hand, if these solutions could not meet the user requirements, the user could easily create or extend the existing classes to meet their needs while taking advantage of all the functionalities this library provides.

%\pagebreak
\subsubsection{Evaluation}

Meta-Trainer can also be used to evaluate the trained models regardless of the toolkit used for their training. 

One of its main features is that in addition to being able to evaluate the models on their test sets, they can also be evaluated on all compatible datasets (e.g., same languages). This feature is particularly relevant for studying continual learning, domain-shift problems, model generalization capabilities, the effects of the catastrophic forgetting problem, etc.

\begin{figure}[ht]
\centering
\subfloat[Evaluate on its test set \label{fig:eval-same}]
        {\includegraphics[width=0.5\columnwidth]{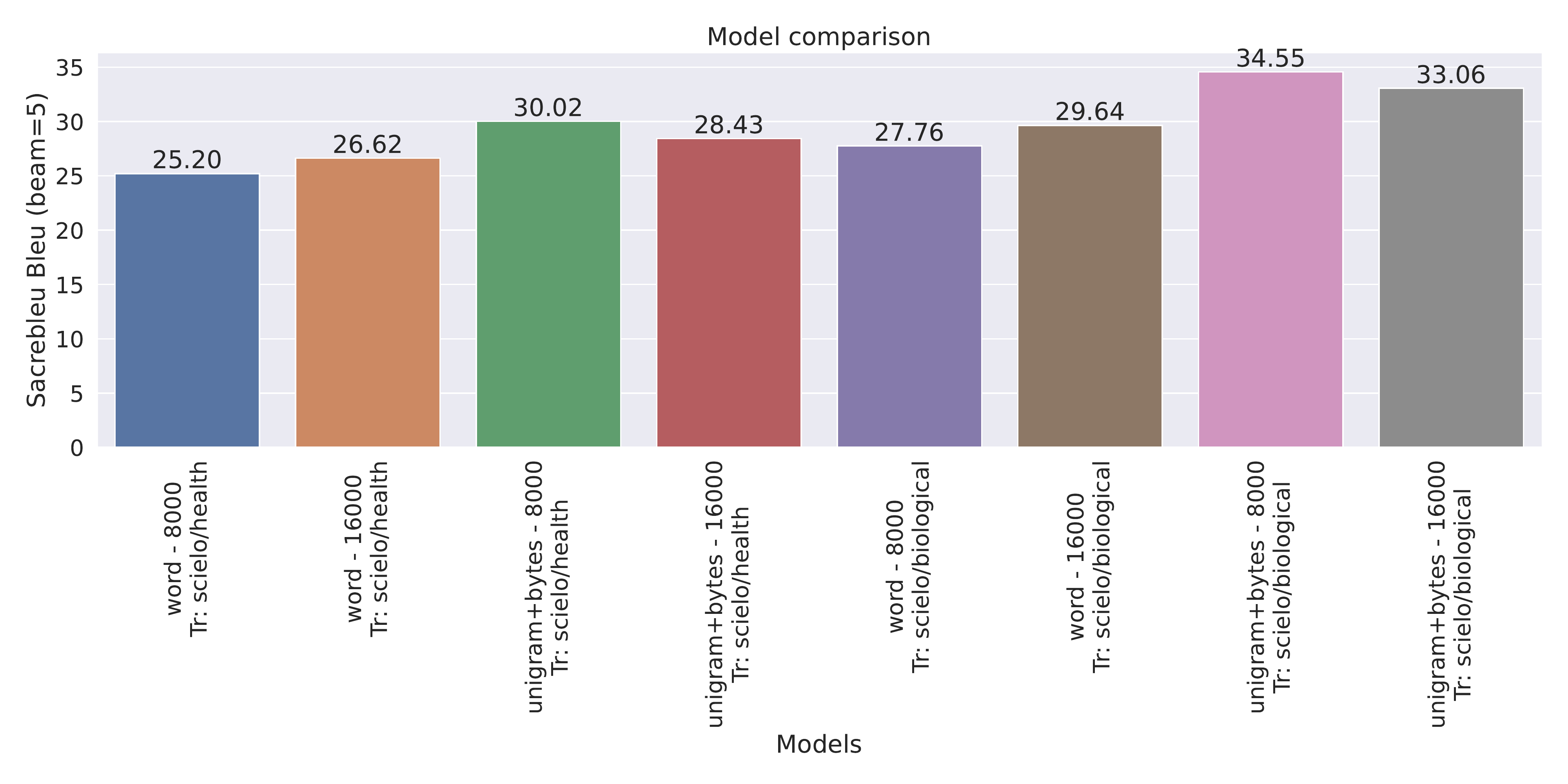}}
%\hfill
\subfloat[Evaluate on compatible datasets \label{fig:eval-compatible}]
     {\includegraphics[width=0.5\columnwidth]{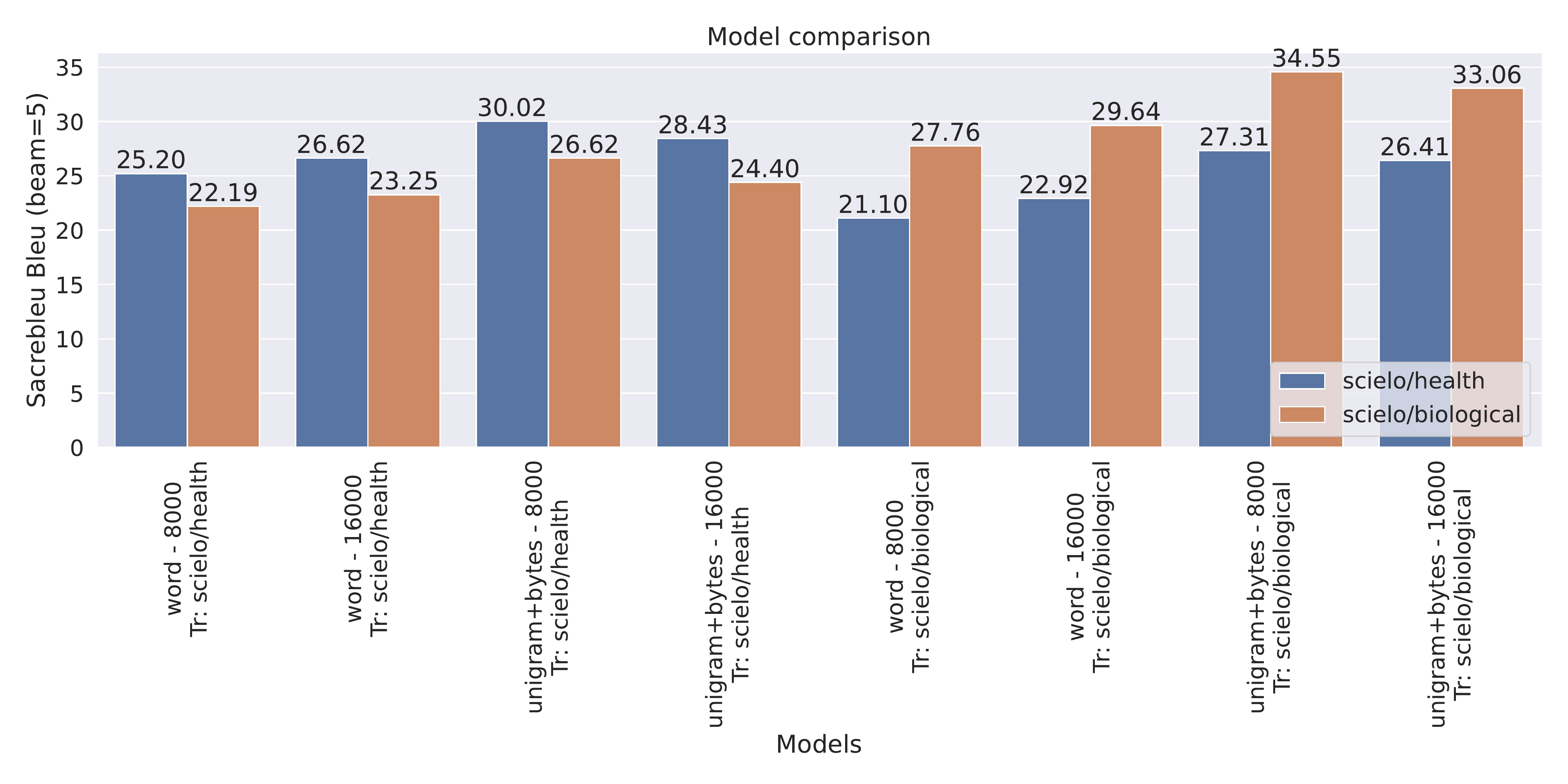}}
\caption{\textbf{Automatic evaluation:} AutoNMT can evaluate each model using the test set of their dataset (See Figure~\ref{fig:eval-same}) or all compatible test sets indexed by the \textit{DatasetBuilder} (See Figure~\ref{fig:eval-compatible}). This feature is particularly useful for studying continual learning or domain-shift problems }
\label{fig:evals}
\end{figure}

Another advantage of this component is that it allows the evaluation of models using lists of arguments with which to specify the metrics to be used (e.g., BLEU, chrF, BERTScore, COMET, BEER, etc.) or the translation settings (e.g., beam width).

\subsection{Reporting}

Creating reports, summaries, and graphs is usually a very time-consuming task. Because of that, we decided to include a set of utilities to collect all available information\footnote{Data and reports are saved as CSV and JSON so that other libraries can load them easily} about the datasets, models, configurations, training, and evaluations to ease its analysis and generate automatic graphs for the following use-cases: 

\begin{itemize}
    \item Evaluate the performance of a model using one or more metrics (See Figure~\ref{fig:eval-same})
    \item Evaluate the generalization capabilities a model (See Figure~\ref{fig:eval-compatible})
    \item Evaluate multiple variables as function of another (See Figure~\ref{fig:case-opt-voc})
\end{itemize}

\section{AutoNMT toolkit}

Given the flexibility of the Meta-Trainer to support multiple toolkits, we decided to write our own toolkit (AutoNMT toolkit) so that users could easily extend it to create new models, tasks, or non-standard trainings (e.g., specific data augmentations of the fly, custom teacher-student approaches, etc.)

As with any other toolkit, the AutoNMT toolkit simply inherits from the \textit{BaseTranslator} class and overrides its default methods (preprocess, train and translate). However, the Trainer defined in this class inherits from the \textit{LightningModule} class, which allows us to create scalable models that can run on distributed hardware seamlessly. 

The main reason for using PyTorch-Lightning was to work within a well-known research framework that lets advanced users modify the code without having to learn the inner workings of our library. A second reason to use PyTorch-Lightning was to boost our Trainer with features such as data parallelization, distributed training, mixed precision, early stopping, logging, fault-tolerant training, etc.

\subsubsection{AutoNMT toolkit: comparison}

In order to demonstrate the competitiveness of our toolkit, we decided to compare it against other reference toolkits such as Fairseq and OpenNMT. To do so, we trained multiple models with different configurations using these toolkits on the following datasets: Multi30K, Europarl (de-en), SciELO (Health), SciELO (Biological), among others.

Even though our toolkit is in active development and lacks some of the training features enabled (by default) on the reference toolkits, the performance was remarkably similar. For example, the average difference in performance from the experiments shown in Table~\ref{tab:toolkit-comparison} was 0.25pts of BLEU. This table contains the results from the models trained using Fairseq and AutoNMT (Toolkit) on the SciELO datasets (Health and Biological), under different preprocessing configurations (two subword models: \textit{Word} and \textit{Unigram+Bytes}, and two vocabulary constraints: 16000 and 8000 words). Similarly, more experiments were performed, but the differences in performance remained consistent between toolkits, datasets, and configurations.

\begin{table}[ht]
\centering
\resizebox{\textwidth}{!}{
\begin{tabular}{@{}lllrrr@{}}
\toprule
\textbf{Train domain} & \textbf{Test domain} & \textbf{Subword model} & \textbf{Vocab. size} & \textbf{Fairseq BLEU} & \textbf{AutoNMT BLEU} \\ \midrule
Health                & Health               & Word                   & 8,000                 & 24.22        & 23.95                 \\
Health                & Health               & Word                   & 16,000                & 25.00                 & 25.36        \\
Biological            & Biological           & Word                   & 8,000                 & 26.61                 & 25.66                 \\
Biological            & Biological           & Word                   & 16,000                & 28.31                 & 27.64                 \\
Health                & Health               & Unigram+Bytes          & 8,000                 & 28.41                 & 29.09        \\
Health                & Health               & Unigram+Bytes          & 16,000                & 26.68                 & 26.82        \\
Biological            & Biological           & Unigram+Bytes          & 8,000                 & 32.78                 & 32.00                 \\
Biological            & Biological           & Unigram+Bytes          & 16,000                & 31.12                 & 30.62                 \\ \bottomrule
\end{tabular}
}
\caption{\textbf{Toolkit comparison}: The performance of AutoNMT is remarkably similar to Fairseq even though AutoNMT was missing some relevant training features that were enabled by default in Fairseq.}
\label{tab:toolkit-comparison}
\end{table}

Concerning the raw speed, it is fast enough to challenge these toolkits competitively for the average researcher, given that most data parallelization modes, scaling strategies, and optimization features are available through the \textit{PyTorch-Lightning} module.

\section{Uses cases}

\subsection{Automating experimentation}

The most common use case for this library is automating the experimentation, from data preprocessing and training to evaluation and reporting. Due to the empirical component of much of the research in the field of machine learning, repeating experiments under different configurations and with multiple datasets is a standard practice to improve the robustness of our findings. 

As an example of this use case, after training the models, a user could simply call the \textit{generate\_report} function to generate an automatic report similar to the one in Figure~\ref{fig:eval-same}.

\subsection{Studying domain-shift effects}

Similar to the previous use case, if AutoNMT detects that each of the models has been evaluated for more than one dataset, it will generate a report that is similar to the one in Figure~\ref{fig:eval-compatible}, which can be used to study the problem of domain adaptation.

\subsection{Finding the optimal vocabulary size}

Sometimes we need to study the performance of our models as a function of one or more variables. To do so, we can make use of the \textit{generate\_multivariable\_report} function, which allows us to plot one or more variables as a function of another.

\begin{figure}[ht]
\centering
\subfloat[Multivariable report \#1 \label{fig:high-resource}]{{
\includegraphics[width=.48\textwidth]{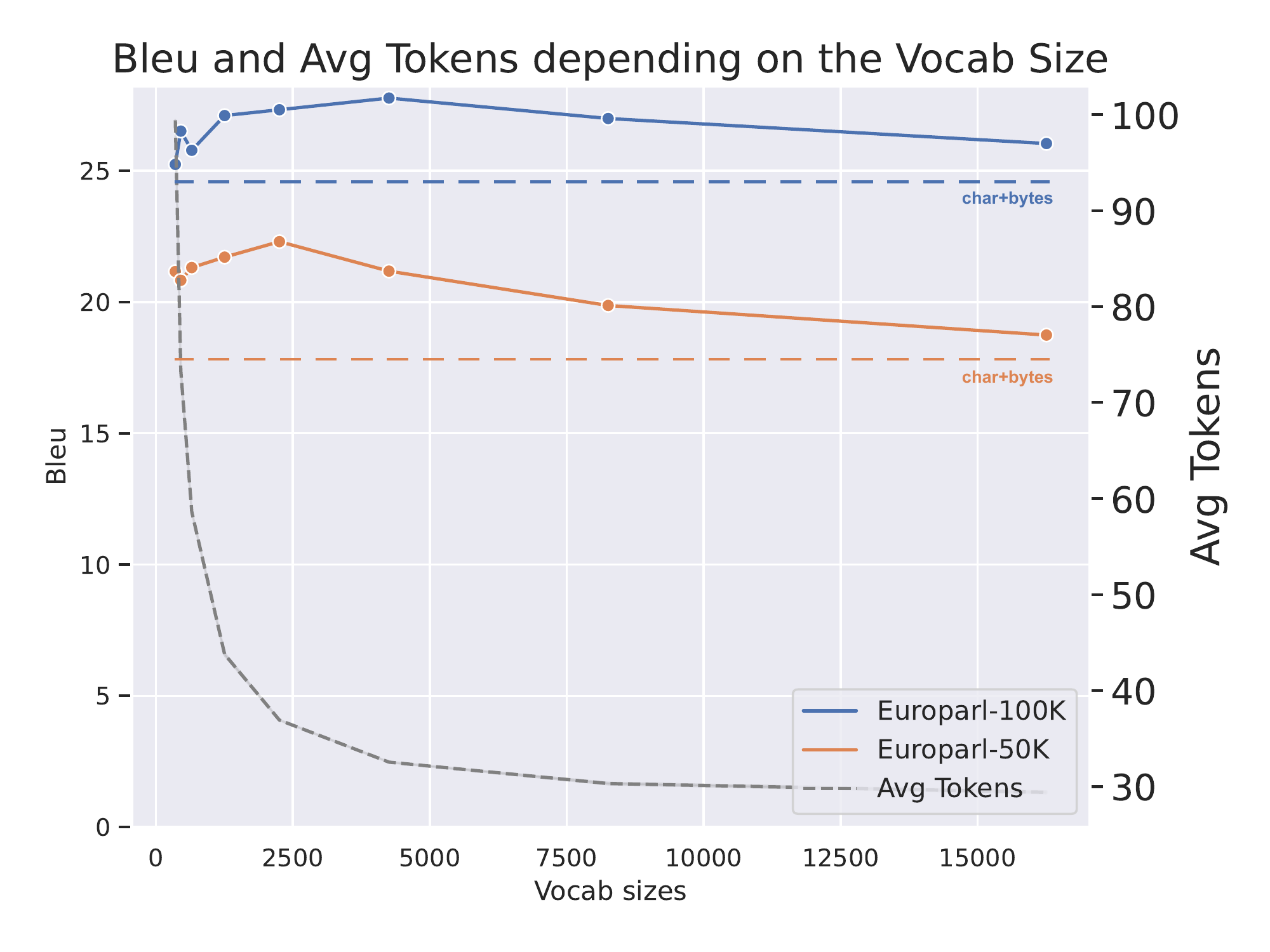}}}
\hfill
\subfloat[Multivariable report \#2 \label{fig:low-resource}]{{
\includegraphics[width=.45\textwidth]{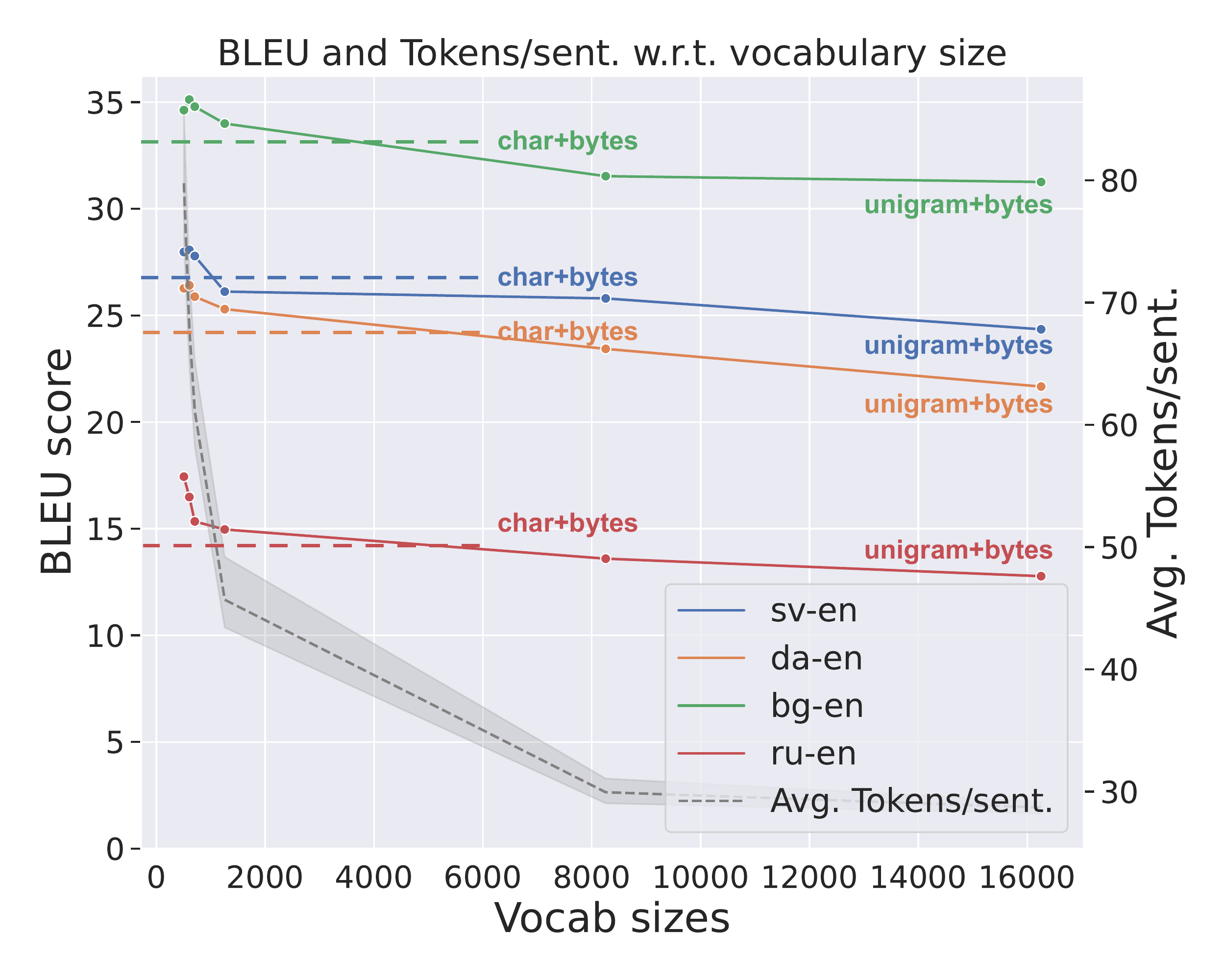}}}
\caption{ \textbf{Multivariable report}: AutoNMT can generate report for one or more variables as a function of another. In this case, we compare the performance of two models and the average number of tokens as a function of the vocabulary size.}
\label{fig:case-opt-voc}
\end{figure}

For example, in Figure~\ref{fig:high-resource} we compare the performance of two models (Europarl-50K and Europarl-100K) measured in BLEU points and the average number of tokens per sentence, as a function of the vocabulary size. Similarly, in Figure~\ref{fig:low-resource}, we compare four models under similar settings.

\subsection{Studying the continual learning problem}

\begin{figure}[htbp]
\centering
\includegraphics[width=0.55\columnwidth]{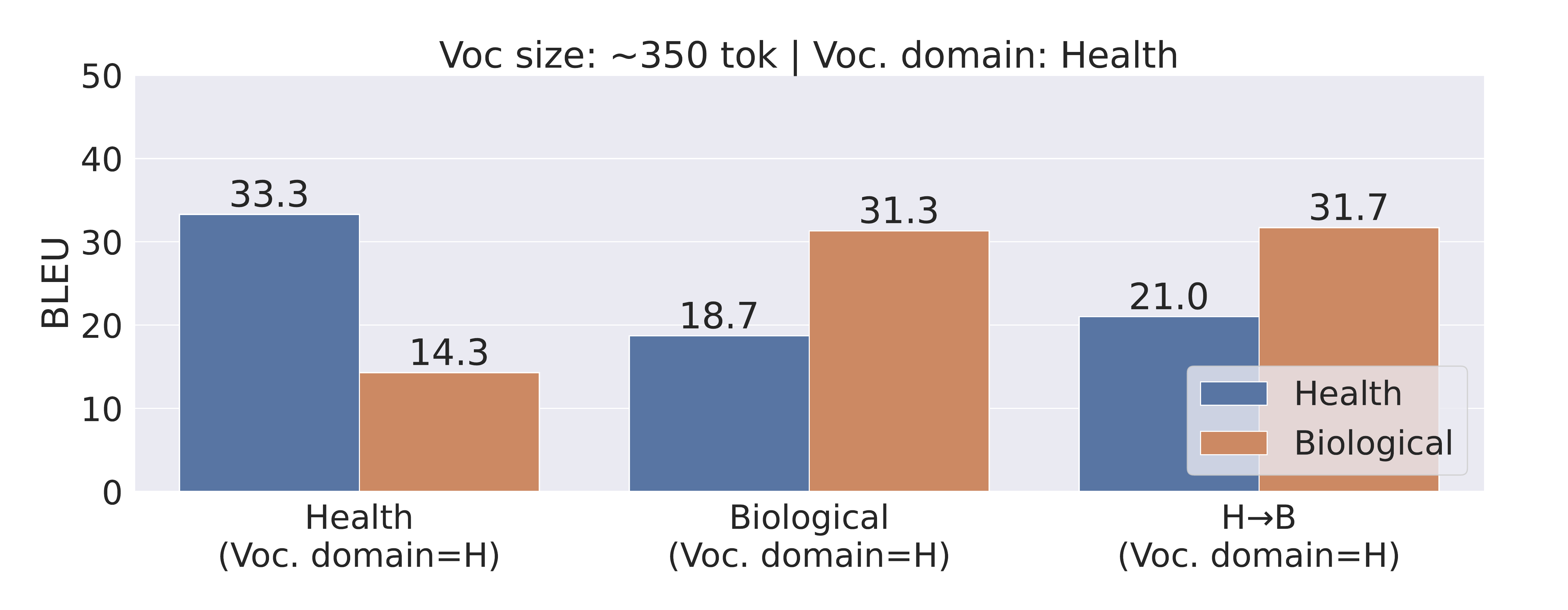}
    \caption{ \textbf{Continual learning report}: AutoNMT can generate reports to visualize the performance of lifelong learning seq-to-seq models, or in this case, the effects of the catastrophic forgetting problem in Neural Machine Translation. }
    \label{fig:cf-problem}
\end{figure}

Finally, we can also use the report generation from AutoNMT to study the continual learning problem in Machine Translation, or the performance of lifelong learning for seq-to-seq models in general. For example, in Figure~\ref{fig:cf-problem} we have generated a report to visualize the effects of the catastrophic forgetting problem for a model trained in the SciELO (Health) dataset.

Another feature typically used in this scenario is filtering line pairs in the dataset by language, domain, or a tag placed at the beginning of the sentence.

\section{Conclusions}

In this paper, we have introduced AutoNMT, a framework to streamline the research of seq-to-seq models through automation and abstraction. 

First, we have presented the three core components of the framework: i) the \textit{DatasetBuilder}, which is the component in charge of automating the data pipeline (file management and preprocessing); ii) the \textit{Meta-Trainer}, which is the component in charge of automating the experimentation (training and evaluation) in a toolkit-agnostic manner; and iii) a utility to generate automatic reports.

Finally, we have presented our seq-to-seq toolkit, the AutoNMT Toolkit, along with a performance comparison against state-of-the-art toolkits and four use cases where this framework can be typically used.

\section{Future Work}

In future work, AutoNMT will support more features, toolkits, and tasks.

\subsubsection*{Acknowledgment}

Work supported by the Horizon 2020 - European Commission (H2020) under the SELENE project (grant agreement no 871467) and the project Deep learning for adaptive and multimodal interaction in pattern recognition (DeepPattern) (grant agreement PROMETEO/2019/121). We gratefully acknowledge the support of NVIDIA Corporation with the donation of a GPU used for part of this research.

\small

\bibliographystyle{apalike}
\bibliography{amta2022}

\begin{thebibliography}{}

\bibitem[Abadi et~al., 2015]{tensorflow}
Abadi, M., Agarwal, A., Barham, P., Brevdo, E., Chen, Z., Citro, C., Corrado,
  G.~S., Davis, A., Dean, J., Devin, M., Ghemawat, S., Goodfellow, I., Harp,
  A., Irving, G., Isard, M., Jia, Y., Jozefowicz, R., Kaiser, L., Kudlur, M.,
  Levenberg, J., Man\'{e}, D., Monga, R., Moore, S., Murray, D., Olah, C.,
  Schuster, M., Shlens, J., Steiner, B., Sutskever, I., Talwar, K., Tucker, P.,
  Vanhoucke, V., Vasudevan, V., Vi\'{e}gas, F., Vinyals, O., Warden, P.,
  Wattenberg, M., Wicke, M., Yu, Y., and Zheng, X. (2015).
\newblock {TensorFlow}: Large-scale machine learning on heterogeneous systems.
\newblock Software available from tensorflow.org.

\bibitem[Bai et~al., 2019]{onnx}
Bai, J., Lu, F., Zhang, K., et~al. (2019).
\newblock Onnx: Open neural network exchange.
\newblock \url{https://github.com/onnx/onnx}.

\bibitem[Chollet et~al., 2015]{keras}
Chollet, F. et~al. (2015).
\newblock Keras.
\newblock \url{https://keras.io}.

\bibitem[Feurer et~al., 2015]{automl}
Feurer, M., Klein, A., Eggensperger, K., Springenberg, J., Blum, M., and
  Hutter, F. (2015).
\newblock Efficient and robust automated machine learning.
\newblock In Cortes, C., Lawrence, N., Lee, D., Sugiyama, M., and Garnett, R.,
  editors, {\em Advances in Neural Information Processing Systems}, volume~28.
  Curran Associates, Inc.

\bibitem[Klein et~al., 2017]{opennmt}
Klein, G., Kim, Y., Deng, Y., Senellart, J., and Rush, A. (2017).
\newblock {O}pen{NMT}: Open-source toolkit for neural machine translation.
\newblock In {\em Proceedings of {ACL} 2017, System Demonstrations}, pages
  67--72, Vancouver, Canada. Association for Computational Linguistics.

\bibitem[Koehn et~al., 2007]{moses}
Koehn, P., Hoang, H., Birch, A., Callison-Burch, C., Federico, M., Bertoldi,
  N., Cowan, B., Shen, W., Moran, C., Zens, R., Dyer, C., Bojar, O.,
  Constantin, A., and Herbst, E. (2007).
\newblock Moses: Open source toolkit for statistical machine translation.
\newblock In {\em Proceedings of the 45th Annual Meeting of the ACL on
  Interactive Poster and Demonstration Sessions}, ACL '07, page 177–180.

\bibitem[Kudo and Richardson, 2018]{sentencepiece}
Kudo, T. and Richardson, J. (2018).
\newblock Sentencepiece: {A} simple and language independent subword tokenizer
  and detokenizer for neural text processing.
\newblock In Blanco, E. and Lu, W., editors, {\em Proceedings of the 2018
  Conference on Empirical Methods in Natural Language Processing, {EMNLP} 2018:
  System Demonstrations, Brussels, Belgium, October 31 - November 4, 2018},
  pages 66--71.

\bibitem[Liaw et~al., 2018]{raytune}
Liaw, R., Liang, E., Nishihara, R., Moritz, P., Gonzalez, J.~E., and Stoica, I.
  (2018).
\newblock Tune: A research platform for distributed model selection and
  training.
\newblock {\em arXiv preprint arXiv:1807.05118}.

\bibitem[Ott et~al., 2019]{fairseq}
Ott, M., Edunov, S., Baevski, A., Fan, A., Gross, S., Ng, N., Grangier, D., and
  Auli, M. (2019).
\newblock fairseq: {A} fast, extensible toolkit for sequence modeling.
\newblock {\em CoRR}, abs/1904.01038.

\bibitem[Papineni et~al., 2002]{metric-bleu}
Papineni, K., Roukos, S., Ward, T., and Zhu, W.-J. (2002).
\newblock Bleu: A method for automatic evaluation of machine translation.
\newblock In {\em Proceedings of the 40th Annual Meeting on ACL}, ACL '02, page
  311–318.

\bibitem[Paszke et~al., 2019]{pytorch}
Paszke, A., Gross, S., Massa, F., Lerer, A., Bradbury, J., Chanan, G., Killeen,
  T., Lin, Z., Gimelshein, N., Antiga, L., Desmaison, A., Kopf, A., Yang, E.,
  DeVito, Z., Raison, M., Tejani, A., Chilamkurthy, S., Steiner, B., Fang, L.,
  Bai, J., and Chintala, S. (2019).
\newblock Pytorch: An imperative style, high-performance deep learning library.
\newblock In Wallach, H., Larochelle, H., Beygelzimer, A., d\textquotesingle
  Alch\'{e}-Buc, F., Fox, E., and Garnett, R., editors, {\em NIPS 32}, pages
  8024--8035.

\bibitem[Pedregosa et~al., 2011]{scikit-learn}
Pedregosa, F., Varoquaux, G., Gramfort, A., Michel, V., Thirion, B., Grisel,
  O., Blondel, M., Prettenhofer, P., Weiss, R., Dubourg, V., Vanderplas, J.,
  Passos, A., Cournapeau, D., Brucher, M., Perrot, M., and Duchesnay, E.
  (2011).
\newblock Scikit-learn: Machine learning in {P}ython.
\newblock {\em Journal of Machine Learning Research}, 12:2825--2830.

\bibitem[Popovi{\'c}, 2015]{metric-chrf}
Popovi{\'c}, M. (2015).
\newblock chr{F}: character n-gram {F}-score for automatic {MT} evaluation.
\newblock In {\em Proceedings of the Tenth WMT}, pages 392--395.

\bibitem[Post, 2018]{sacrebleu}
Post, M. (2018).
\newblock A call for clarity in reporting {BLEU} scores.
\newblock In {\em Proceedings of the Third Conference on Machine Translation:
  Research Papers}, pages 186--191.

\bibitem[Rei et~al., 2020]{metric-comet}
Rei, R., Stewart, C., Farinha, A.~C., and Lavie, A. (2020).
\newblock {COMET:} {A} neural framework for {MT} evaluation.
\newblock {\em CoRR}, abs/2009.09025.

\bibitem[Stanojevi{\'c} and Sima{'}an, 2014]{metric-beer}
Stanojevi{\'c}, M. and Sima{'}an, K. (2014).
\newblock Fitting sentence level translation evaluation with many dense
  features.
\newblock In {\em Proceedings of the 2014 Conference on Empirical Methods in
  Natural Language Processing ({EMNLP})}, pages 202--206, Doha, Qatar.
  Association for Computational Linguistics.

\bibitem[Wolf et~al., 2019]{huggingface}
Wolf, T., Debut, L., Sanh, V., Chaumond, J., Delangue, C., Moi, A., Cistac, P.,
  Rault, T., Louf, R., Funtowicz, M., and Brew, J. (2019).
\newblock Huggingface's transformers: State-of-the-art natural language
  processing.
\newblock {\em CoRR}, abs/1910.03771.

\bibitem[Zhang et~al., 2019]{metric-bertscore}
Zhang, T., Kishore, V., Wu, F., Weinberger, K.~Q., and Artzi, Y. (2019).
\newblock Bertscore: Evaluating text generation with {BERT}.
\newblock {\em CoRR}, abs/1904.09675.

\end{thebibliography}

\end{document}